\documentclass[a4paper,french]{rnti}
\usepackage{amssymb}
\usepackage{amsfonts}
\usepackage{graphicx}
\usepackage{algorithmic}
\usepackage{algorithm}
\usepackage{url}
\usepackage[latin1]{inputenc}

\floatname{algorithm}{algorithme}

\newcommand{\R}[1]{\ensuremath{\mathbb{R}^{#1}}}

\newcommand{\vdata}{\mathbf{x}}
\newcommand{\vmodel}{\mathbf{m}}
\newcommand{\vmodelfull}{\mathcal{M}}
\newcommand{\graph}{\mathcal{G}}
\newcommand{\vertices}{\Gamma}
\newcommand{\edges}{\mathcal{A}}
\newcommand{\graphdist}{\delta}
\newcommand{\neigh}{h}
\newcommand{\kernel}{K}
\newcommand{\affect}{c}

\newcommand{\cluster}{\mathcal{C}}
\newcommand{\diss}{d}
\newcommand{\inputspace}{\mathcal{X}}
\newcommand{\inputset}{\Omega}
\newcommand{\energy}{\mathcal{E}}
\newcommand{\partition}{\mathcal{P}}
\newcommand{\minorant}{\zeta}
\newcommand{\minorantfull}{\zeta}
\newcommand{\realminD}{\lambda}
\newcommand{\wsum}{S}
\newcommand{\verticessubset}{\gamma}

\newcommand{\REM}[1]{}

\titrecourt{Heuristiques d'accélération pour le DSOM}

\nomcourt{B. Conan-Guez et F. Rossi}

\titre{Accélération des cartes auto-organisatrices sur tableau de
  dissimilarités par séparation et évaluation} 

\auteur{Brieuc Conan-Guez\affil{1},
        Fabrice Rossi\affil{2}}

\affiliation{
    \affil{1}LITA EA3097, Universit\'e de Metz, Ile du Saulcy, F-57045
  Metz\\
  Brieuc.Conan-Guez@univ-metz.fr\\

    \affil{2}Projet AxIS, INRIA, Domaine de Voluceau, Rocquencourt,
  B.P. 105, \\ 78153 Le Chesnay Cedex\\
          Fabrice.Rossi@inria.fr
 }

\resume{%
Dans cet article, nous proposons une nouvelle implémentation d'une adaptation
des cartes auto-organisatrices de Kohonen (SOM) aux tableaux de dissimilarités.
Cette implémentation s'appuie sur le principe de séparation et évaluation afin 
de réduire le temps de calcul global de l'algorithme. Une propriété
importante de ce nouvel algorithme tient au fait que les résultats produits
sont strictement identiques à ceux de l'algorithme original. 
}

\summary{%
In this paper, a new implementation of the adaptation of Kohonen self-organising maps (SOM)
to dissimilarity matrices is proposed. This implementation relies on the branch and bound principle
to reduce the algorithm running time. An important property of this new approach
is that the obtained algorithm produces exactly the same results as the standard algorithm. 
 
}
\begin{document}
%-------------------------------------------------------------------------------------------------
\section{Introduction}

Dans beaucoup d'applications réelles, les individus étudiés ne peuvent pas
être décrits efficacement par des vecteurs numériques : on pense par exemple à
des données de tailles variables, comme les séquences d'acides aminés
constituant des protéines, ou bien à des données (semi-)structurées (par
exemple des documents XML).  Une solution pour traiter de telles données est
de s'appuyer sur une mesure de dissimilarité permettant de comparer les
individus deux à deux.

Nous nous intéressons dans cet article à une adaptation des cartes
auto-organisatrices de Kohonen (SOM pour {\it Self-Organizing Map},
\citep{KohonenSOM1995}) aux données décrites seulement par un tableau de
dissimilarités, proposée dans
\citep{KohonenSomervuo1998Symbol,KohonenSomervuo2002NonVectorial}.  Nous
désignons cette adaptation par le sigle DSOM (pour \emph{Dissimilarity SOM}).
On trouve aussi dans la littérature l'appellation \emph{Median SOM}. Le DSOM
et ses variantes ont été appliqués avec succès à des problèmes réels d'analyse
exploratoire portant sur des protéines, des données météorologiques
\citep{ElGolliConanGuezRossi04JSDA}, spectrométriques
\citep{ElGolliConanGuezRossiIFCS2004SomDiss} ou encore provenant de l'usage
d'un site web \citep{RossiEtAlESANN2005WUMSOM,ElGolliEtAl06DSOM}. Comme
dans le cas classique, les résultats obtenus par le DSOM
en terme de la qualité de la classification sont comparables à ceux obtenus
avec d'autres méthodes applicables à des tableaux de dissimilarités (comme PAM
\citep{KaufmanRousseeuw1987PAM} ou les algorithmes de type nuées dynamiques
\citep{CeleuxEtAl1989}). L'avantage du DSOM sur ces algorithmes réside dans
prise en compte d'une structure \emph{a priori} qui permet la représentation
graphique des classes et prototypes obtenus, ce qui facilite l'analyse
exploratoire des données étudiées. 

Le problème majeur du DSOM réside dans son temps de calcul : pour $N$
observations et $M$ classes, le coût algorithmique du DSOM est de
$O(N^2M+NM^2)$ par itération. A titre de comparaison, un algorithme comme PAM
\citep{KaufmanRousseeuw1987PAM} a un coût de $O(N^2)$ par itération, ce qui le
rend utilisable pour des données beaucoup plus volumineuses. Dans nos
travaux précédents
\citep{ConanGuezEtAlWSOM2005,ConanGuezEtAl06FastDSOM,ConanGuezEtAl06FastDSOMSFC},
nous avons proposé une modification algorithmique (rappelée à la section
\ref{sectionPartialSum} du présent article) qui ramène ce coût à
$O(N^2+NM^2)$, tout en conservant exactement les mêmes résultats. La réduction
du coût théorique a été confirmée par des expériences montrant une très forte
diminution du temps de calcul. Nous avons en outre étudié diverses
heuristiques d'accélération des calculs qui permettaient de gagner un facteur
allant jusqu'à 3.5 par rapport au code en $O(N^2+NM^2)$, tout en garantissant
encore des résultats identiques à ceux de l'algorithme d'origine.

Dans le présent article, nous proposons de nouvelles modifications de
l'algorithme, basées sur le principe de séparation et évaluation (\emph{branch
  and bound}, \citep{LandDoig1960}). Celui-ci s'appuie sur le calcul d'un
minorant (dans la phase d'évaluation). La qualité de ce minorant ainsi que le
temps nécessaire pour le calculer influent de manière importante sur le temps
global d'exécution. Nous proposons donc diverses heuristiques de calcul de ce
minorant, et nous illustrons grâce à des expériences menées sur des données
simulées et réelles l'intérêt de ces différentes approches.  Enfin, nous
combinons au principe de séparation et évaluation certaines des techniques
d'implémentations employées dans notre précédent travail, afin de réduire
encore plus le temps de calcul. Nous obtenons ainsi une implémentation dont
les temps de calcul sont comparables à ceux des méthodes concurrentes au DSOM.
Comme pour notre travail antérieur, les résultats produits par cette nouvelle
version du DSOM sont strictement identiques à ceux du DSOM original. Bien que
cet article ne porte que sur le DSOM, la plupart des techniques proposées
peuvent aisément être appliquées à d'autres algorithmes de classification de
tableaux de dissimilarités s'appuyant sur la notion de prototype, en
particulier le {\it Median Neural Gas} proposé dans
\citep{Cottrell06NeuralGas}.
 
%-------------------------------------------------------------------------------------------------
\section{Cartes auto-organisatrices de Kohonen adaptées aux tableaux de dissimilarités}
Nous rappelons dans cette section l'adaptation du SOM aux tableaux de
dissimilarités (DSOM,
\citep{KohonenSomervuo1998Symbol,KohonenSomervuo2002NonVectorial,ElGolliConanGuezRossi04JSDA}).
On considère $N$ individus $\inputset=(\vdata_i)_{1 \leq i \leq N}$
appartenant à un espace arbitraire $\inputspace$. Cet espace est muni d'une
mesure de dissimilarité $\diss$, qui vérifie les propriétés classiques :
symétrie ($\diss(\vdata_i,\vdata_j)=\diss(\vdata_j,\vdata_i)$), positivité
($\diss(\vdata_i,\vdata_j)\geq 0$), et self-égalité
($\diss(\vdata_i,\vdata_i)=0$).

Comme pour le SOM, le but du DSOM est double : construire un ensemble de
prototypes représentatifs des données et organisés selon une structure \emph{a
  priori}. Cette dernière est décrite par un graphe non orienté
$\graph=(\vertices,\edges)$, où $\vertices$ est l'ensemble des n{\oe}uds (ou
sommets) du graphe, numérotés de $1$ à $M$ ($\vertices=\{1,\ldots,M\}$), et
$\edges$ est l'ensemble des arêtes. On munit $\graph$ d'une distance, notée $\graphdist$. 
A l'instar du SOM, le DSOM est souvent utilisé
afin d'obtenir une visualisation simplifiée d'un jeu de données. Dans ce cas,
il est usuel de définir le graphe $\graph$ comme un maillage régulier (rectangulaire ou
hexagonal) appartenant à un espace bidimensionnel: chaque n{\oe}ud est un élément
du plan. La distance de graphe $\graphdist$ est alors définie comme étant
soit la distance de graphe naturelle (la distance séparant deux n{\oe}uds
est donnée par le nombre d'arêtes composant le chemin de longueur minimale 
entre les deux n{\oe}uds considérés), 
soit la distance euclidienne (distance dans le plan entre les deux n{\oe}uds). 

A chaque noeud du graphe, on associe un prototype $\vmodel^l_j$ qui est choisi
dans $\inputset$ (pour chaque n{\oe}ud $j$, il existe $i$ tel que
$\vmodel^l_j=\vdata_i$). L'exposant $l$ indique que le prototype est celui
obtenu à l'itération $l$ de l'algorithme. On voit apparaître dans le choix des
prototypes une des différences majeures entre l'algorithme du SOM standard et
le DSOM : pour le SOM standard, les prototypes sont décrits par des vecteurs
quelconques choisis sans contrainte dans $\inputspace$. Pour le DSOM en
revanche, il n'est pas possible de construire des éléments arbitraires de
$\inputspace$ et on impose donc aux prototypes d'être choisis parmi les
données.

La quantification réalisée par les prototypes est représentée par une fonction
d'affectation $\affect^l(i)$, qui à chaque individu $\vdata_i$ associe un
n{\oe}ud du graphe : le prototype associé au noeud, $\vmodel^l_{\affect^l(i)}$
est le représentant de l'individu concerné. On note $\cluster_j^l$ la classe
associée au n{\oe}ud $j$ à l'itération $l$ : $\cluster_j^l=\{ \vdata_i \in \inputset |
\affect^l(i)=j\}$.  Les $\cluster_j^l$ forment une partition $\partition^l$ de
$\inputset$.

La structure de graphe impose une contrainte topologique par l'intermédiaire
d'une fonction de voisinage définie à partir d'une fonction noyau décroissante
$\kernel$ (par exemple $\kernel(x)=\exp(-x^2)$) de la façon suivante :
$\neigh(j,k)=\kernel(\graphdist(j,k))$. Le but de la fonction de voisinage est
de mesurer l'influence d'un n{\oe}ud $j$ du graphe sur ses voisins : deux
prototypes associés à des n{\oe}uds voisins sur le graphe auront des
comportements relativement identiques, alors que deux prototypes associés à
des n{\oe}uds distants sur le graphe seront libres d'évoluer indépendemment.

La fonction de voisinage évolue à chaque itération $l$ : on note
$\neigh^l(.,.)$ la fonction utilisée à l'itération $l$ (on l'obtient
généralement par une relation de la forme
$\neigh(j,k)=\kernel\left(\frac{\graphdist(j,k)}{T^l}\right)$, où $T^l$
diminue avec $l$). La suite de fonctions $(\neigh^l)_l$ vérifie dans la
pratique deux propriétés importantes. $(\neigh^l)_l$ est décroissante (i.e.,
$\neigh^l>\neigh^{l'}$ si $l<l'$). De plus, lors des dernières itérations $l$
de l'algorithme, $\neigh^l$ peut être assimilée à une fonction delta de
Kronecker. Ces deux propriétés font que la contrainte topologique se relâche
au cours des itérations, l'algorithme se comportant en fin d'apprentissage
comme un algorithme de type nuées dynamiques.

L'algorithme du DSOM, qui est basé sur la version non stochastique du SOM 
(version \emph{batch}), cherche à minimiser la fonction énergie suivante :
\begin{equation}\label{eq:Energy}
\energy^l(\partition^l,\vmodelfull^l)=\sum_{i=1}^N \sum_{j \in
  \vertices}\neigh^l(\affect^l(i),j)  \diss(\vdata_i,\vmodel_j^l).
\end{equation}
Cette fonction mesure l'adéquation entre une partition des individus
$\partition^l$ et un ensemble de prototypes $\vmodelfull^l$, en tenant compte
de la topologie. L'algorithme commence par une \emph{phase d'initialisation} :
on choisit par exemple $\vmodelfull^0=(\vmodel_1^0,\ldots,\vmodel_M^0)$ de
manière aléatoire.  Puis l'algorithme alterne les \emph{phases d'affectation}
et les \emph{phases de représentation} jusqu'à la convergence. Pour
l'itération $l$, on a :
\begin{enumerate}
\item \emph{phase d'affectation} : chaque individu $\vdata_i$ est affecté au
  prototype qui le représente le mieux. Pour ce faire, on peut choisir le
  prototype le plus proche et donc poser $\affect^l(i)=\arg \min_{j \in
    \vertices} \diss(\vdata_i,\vmodel_j^{l-1})$.  Dans la pratique cependant,
  cette règle d'affectation, bien que classique, présente un inconvénient
  important : elle ne permet pas de lever les ambiguïtés relatives aux
  collisions entre prototypes (cas relativement fréquent où deux n{\oe}uds
  distincts sont associés au même prototype). Pour contourner ce problème,
  nous utilisons la modification proposée dans
  \citep{KohonenSomervuo1998Symbol}. Cette méthode consiste à affiner le
  critère d'affectation dans le cas où une collision entre prototypes se
  produit : on ne cherche plus à trouver le prototype le plus proche de
  l'individu, mais le prototype dont l'ensemble des voisins est le plus proche
  de l'individu (le rayon de recherche sur le graphe est incrémenté de 1 tant
  qu'un prototype n'est pas déclaré vainqueur).  Remarquons que ces deux
  règles d'affectation ne garantissent pas la décroissance de la fonction
  énergie à chaque itération : on observe cependant dans la pratique la
  décroissance en moyenne de celle-ci. On pourrait aussi affecter les
  individus en minimisant directement $\energy^l(\partition^l,\vmodelfull^l)$
  par rapport à $\partition^l$, en considérant $\vmodelfull^l$ fixé, comme
  dans \citep{ElGolliConanGuezRossiIFCS2004SomDiss}. Cela conduit en général à
  des résultats similaires à ceux obtenus avec la méthode retenue ici, mais au
  prix d'une légère augmentation du temps de calcul. 
\item \emph{phase de représentation} : l'algorithme cherche de nouvelles
  valeurs pour les prototypes (i.e., $\vmodelfull^l$) en minimisant la fonction
  énergie $\energy^l(\partition^l,.)$ pour une partition $\partition^l$ fixée.
  Ce problème d'optimisation peut se réécrire facilement comme la somme de $M$
  problèmes d'optimisation indépendants (un problème par n{\oe}ud du graphe).
  Chaque prototype $\vmodel_j^l$ est donc solution du nouveau problème :
\begin{equation}
\label{equRepresentation}
\vmodel_j^l = \arg \min_{\vmodel \in \inputset} \sum_{i=1}^N \neigh^l(\affect^l(i),j) \diss(\vdata_i,\vmodel).
\end{equation}
\end{enumerate}

%-------------------------------------------------------------------------------------------------
\section{Recherche exhaustive efficace}\label{sectionPartialSum}
Si l'on examine le coût algorithmique du DSOM, on constate que pour une
itération de l'algorithme, le coût de la phase d'affectation est en $O(NM^2)$
au pire 
(voir la règle d'affectation modifiée dans \citep{KohonenSomervuo1998Symbol}) et que la
phase de représentation (voir équation \ref{equRepresentation}) est en
$O(N^2M)$\footnote{à comparer avec $O(nNM)$ dans le cas du SOM \emph{batch}
  classique sur des vecteurs de dimension $n$.} dans le cadre d'une recherche
exhaustive. Comme $N>M$ dans tous les cas, la phase de représentation domine
nettement le calcul. 

Cependant, nous avons montré
\citep{ConanGuezEtAlWSOM2005,ConanGuezEtAl06FastDSOM,ConanGuezEtAl06FastDSOMSFC}
qu'il était possible d'exploiter la structure de l'équation
\ref{equRepresentation} pour réduire le coût d'une recherche exhaustive. A
l'itération $l$ et pour chaque n{\oe}ud $j$, on cherche à trouver pour quel
$k$, la fonction $\wsum^l(j,k)$ est minimale, où $\wsum^l(j,k)=\sum_{i=1}^N
\neigh^l(\affect^l(i),j) \diss(\vdata_i,\vdata_k)$.  Si l'on note
$D^l(u,k)=\sum_{\vdata_i \in \cluster_u^l} \diss(\vdata_i,\vdata_k)$, on peut
exprimer $\wsum^l(j,k)$ en regroupant les individus par classes :
\begin{equation}
\label{equPartialSum}
\wsum^l(j,k)=\sum_{u=1}^M \neigh^l(u,j) \sum_{\vdata_i \in \cluster_u^l} \diss(\vdata_i,\vdata_k)=\sum_{u=1}^M \neigh^l(u,j) D^l(u,k).
\end{equation}
Il y a $MN$ différentes valeurs $D^l(u,k)$ (nommées \textbf{sommes partielles}
dans la suite de l'article), qui peuvent être pré-calculées une fois pour
toute avant la phase de représentation. On montre dans
\citep{ConanGuezEtAl06FastDSOM} que le coût de cette phase de pré-calcul est
en $O(N^2)$ et que celui de la recherche exhaustive s'appuyant sur cette
nouvelle formulation est en $O(NM^2)$. On a donc un coût total pour la phase
de représentation de $O(N^2+NM^2)$, à comparer avec $O(N^2M)$ pour
l'algorithme initial. Comme on a $N>M$ dans toutes les situations, cette
approche réduit le coût du DSOM. De plus, les résultats obtenus sont
strictement identiques à ceux de l'algorithme initial.

\section{Accélération par séparation et évaluation}
Dans la section précédente, les modifications apportées à l'algorithme du DSOM
ont permis de réduire sa complexité algorithmique de manière significative.
Dans cette section, nous allons exploiter le principe général de séparation et
évaluation (\emph{branch and bound}) pour rendre encore plus efficace la phase
de représentation en évitant une recherche exhaustive naïve.

\subsection{Séparation et évaluation}
Rappelons tout d'abord le principe général de séparation et évaluation. Quand
on doit minimiser une fonction $f$ sur un domaine $K$, on cherche à éviter
l'exploration exhaustive de $K$. Pour ce faire, on se donne une décomposition
de $K$ en sous-ensembles (si possible disjoints) qui correspond au terme
\emph{séparation}. On suppose être capable d'évaluer efficacement un minorant
des valeurs de $f$ sur chaque sous-ensemble (ce qui correspond au terme
\emph{évaluation}). Au lieu de faire une recherche exhaustive dans $K$, on
commence par une région donnée, dans laquelle la recherche est conduite de
façon exhaustive, ce qui conduit à un majorant pour le minimum. On compare
alors ce majorant au minorant d'une région non explorée jusqu'à présent. Si le
minorant est plus grand que le majorant actuel, on peut éliminer la région
entière sans avoir besoin de l'explorer. Dans le cas contraire, on met à jour
le majorant grâce à l'exploration exhaustive de la région.

Les performances d'un algorithme exploitant la séparation et l'évaluation
dépendent de plusieurs facteurs : la rapidité de calcul des minorants dans une
région, la qualité de ces minorants, l'ordre de parcours des régions, etc. 

\subsection{Séparation pour  la représentation}
Dans la phase de représentation du n{\oe}ud $j$, on cherche le minimum sur
$\inputset$ (identifié à $\{1,\ldots,N\}$) de $\wsum^l(j,.)$. Nous proposons
de réaliser la séparation de $\inputset$ à partir de la partition produite par
l'algorithme lors de la phase d'affectation ($\partition^l$). Le but du DSOM
est en effet de produire des classes homogènes, mais aussi séparées (bien que
la séparation entre classes ne soit pas maximisée explicitement). En raison de
l'homogénéité, la valeur de $\wsum^l(j,.)$ reste relativement constante sur
une classe et un minorant est donc assez représentatif des valeurs attendues.
En raison de la séparation, les éléments d'une classe $i$ ne devraient pas
être de bons candidats pour représenter la classe $j\neq i$ : on peut donc
espérer qu'un minorant de $\wsum^l(j,.)$ pour les éléments de $\cluster_i^l$
soit relativement élevé et rende donc superflue une recherche exhaustive dans
cette classe (nous reviendrons sur ce point dans la section
\ref{subSectionOrdre}).

Si l'on fait l'hypothèse d'une équi-répartition des observations dans les
classes (soit donc $\frac{N}{M}$ individus par classe), l'approche séparation
et évaluation peut réduire considérablement le coût de la phase de
représentation. Considérons en effet le cas d'un n{\oe}ud fixé, pour lequel la
recherche exhaustive est en $O(NM)$. Dans le cas le plus favorable, on
n'effectue une recherche exhaustive seulement dans la première classe
considérée. Pour chaque individu de cette classe, le coût d'évaluation de
$\wsum^l(j,k)$ est de $O(M)$, soit donc un coût total en $O(N)$ (avec
équi-répartition). Pour les $M-1$ classes restantes, on se contente de
comparer l'évaluation du minorant au majorant obtenu grâce à la première
classe, ce qui ajoute un temps de calcul en $O(M)$ (en ne tenant pas compte de
la phase d'évaluation pour l'instant). 

Dans le cas idéal, on passe donc d'une phase de représentation en $O(NM^2)$ à
une phase en $O(M(N+M))$, à laquelle on doit ajouter le temps de calcul des
évaluations (complexité en $O(MN)$).

\subsection{Ordre de parcours des classes}\label{subSectionOrdre}
L'efficacité de l'approche séparation et évaluation est fortement liée à
l'ordre dans lequel les groupes d'éléments sont parcourus : l'algorithme est
d'autant plus efficace qu'on trouve rapidement un bon majorant du minimum. Or,
la partition calculée par le DSOM ordonne les données en un sens fortement lié
au problème d'optimisation de la phase de représentation.

En effet, les individus affectés à une classe $\cluster_j^l$ sont par
définition proches (au sens de la dissimilarité) du prototype
$\vmodel^{l-1}_j$ de cette classe. Il semble donc naturel de chercher la
valeur de $\vmodel^{l}_j$ d'abord dans les éléments de $\cluster_j^l$. Comme
nous le verrons à la section \ref{sectionExperiences}, cette stratégie s'avère
en pratique très efficace. 

En raison de la structure \emph{a priori} imposée aux prototypes par
l'algorithme du DSOM, le prototype d'un n{\oe}ud $j$ est aussi un modèle
correct pour les éléments affectés aux n{\oe}uds voisins de $j$ dans le
graphe. De ce fait, ces éléments sont des candidats potentiels intéressant
pour trouver le prototype de $j$. Une fois qu'on a parcouru de façon
exhaustive $\cluster_j^l$, il semble donc pertinent de passer aux classes
voisines, i.e., aux $\cluster_u^l$ pour lesquelles $\graphdist(j,u)$ est
petit. En pratique cependant, cette solution n'améliore que marginalement les
résultats et nous ne l'avons pas incluse dans les expériences proposées dans
l'article. En effet, après quelques itérations de l'algorithme, les classes
sont déjà suffisamment séparées pour que le prototype d'une classe soit très
fréquemment un élément de celle-ci. L'évaluation produit alors généralement un
minorant pour les autres classes supérieur à la valeur de $\wsum^l(j,.)$ pour
le candidat trouvé dans $\cluster_j^l$, ce qui évite leur parcours exhaustif.

\subsection{Évaluation pour  la représentation}\label{subsectionEvaluation}
L'évaluation consiste donc à minorer $\wsum^l(j,.)$ sur chaque classe
$\cluster_u^l$ de $\partition^l$. Plus précisément on cherche à approcher par
les valeurs inférieures la valeur $\min_{\vdata_k \in \cluster_u^l} \wsum^l(j,k)$.

La stratégie retenue consiste à exploiter la formulation de l'équation
\ref{equPartialSum} et la positivité des éléments qui apparaissent dans les
$\wsum^l(j,k)$. On constate en effet que

% nouvelle version brieuc
\begin{equation}\label{eqPartialSumMinFull}
\minorantfull^l(j,u,\verticessubset)=\sum_{v \in \verticessubset} \neigh^l(v,j) \min_{\vdata_k \in \cluster_u^l}D^l(v,k)\leq
\min_{\vdata_k \in \cluster_u^l} \wsum^l(j,k).
\end{equation}
où $\verticessubset$ est sous-ensemble quelconque de $\vertices=\{1,\ldots,M\}$.

Le calcul de $\realminD^l(v,u)=\min_{\vdata_k \in \cluster_u^l}D^l(v,k)$
se fait en $O(|\cluster_u^l|)$, et, globalement, celui de l'ensemble des
$\realminD^l(.,.)$ se fait en $O(NM)$. Ceci est compatible avec les plus
faibles temps de calcul envisageables pour la phase de représentation.
Etudions à présent les deux stratégies extrêmes pour le choix du sous-ensemble $\verticessubset$,
i.e. $\verticessubset$ égale à $\vertices$ et  $\verticessubset$ réduit à un singleton:

La première stratégie revient à choisir $\verticessubset = \vertices$.
Dans ce cas, par positivité des éléments sommés, les minorants $\minorantfull^l(j,u,\vertices)$ obtenus sont
maximaux: tout autre choix pour $\verticessubset$ mène à des minorants plus faibles, et donc de moins bonne qualité.
Ce choix permet donc de minimiser le nombre de recherches exhaustives dans les classes pour l'algorithme
du \emph{branch and bound}. Malheureusement, le coût de calcul du minorant de l'équation
\ref{eqPartialSumMinFull} avec $\verticessubset = \vertices$ est alors de $O(M)$ 
pour chaque couple de n{\oe}uds $(j,u)$,
ce qui induit un coût total en $O(M^3)$ pour obtenir l'ensemble des
$\minorantfull^l(.,.,\vertices)$. Le sur-coût induit par cette méthode d'évaluation des
classes est inférieur au coût de la recherche exhaustive (en $O(NM^2)$), mais
largement supérieur à celui du coût minimal envisageable dans le cas d'un
fonctionnement idéal du \emph{branch and bound} ($O(M(N+M))$).
C'est la raison pour laquelle nous envisageons l'alternative suivante. 

La deuxième stratégie revient à restreindre $\verticessubset$ à un singleton.
On obtient donc une borne de moins bonne qualité, mais avec un coût de calcul plus faible, au
total en $O(M^2)$ pour l'ensemble des $\minorant^l$, du même ordre de grandeur
que le coût minimal envisageable (en tenant compte en outre du temps de calcul
des $\realminD^l(.,.)$).

En pratique, nous nous sommes intéressés au cas de
$\minorant^l(j,u,\{j\})$ quand $\verticessubset$ est réduit à un singleton. 
Cette heuristique est motivée par la remarque
suivante. La valeur $\neigh^l(v,j)$ est d'autant plus petite que
la distance $\graphdist(v,j)$ est grande. En outre les fonctions généralement
utilisées pour calculer $\neigh^l(v,j)$ à partir de $\graphdist(v,j)$
décroissent rapidement. En pratique, la valeur $\minorantfull^l(j,u,\{j\})$ est donc
largement influencée par celle de $\realminD^l(j,u)$. 

Nous verrons en pratique à la section \ref{sectionExperiences} que cette
heuristique donne des résultats très intéressants.

\subsection{Algorithme}
Pour le calcul de $\vmodel^l_j$, nous obtenons au final l'algorithme
\ref{algoBranchAndBound} qui s'appuie sur le pré-calcul des grandeurs
suivantes : 
\begin{itemize}
\item les sommes partielles $D^l(u,k)=\sum_{\vdata_i \in \cluster_u^l}
  \diss(\vdata_i,\vdata_k)$ (cf section \ref{sectionPartialSum}) ;
\item les minima, classe par classe, de ces sommes partielles, c'est-à-dire les
  $\realminD^l(v,u)=\min_{\vdata_k \in \cluster_u^l}D^l(v,k)$ (cf section
  \ref{subsectionEvaluation}).
\end{itemize}

\begin{algorithm}[htbp]
\caption{Calcul de $\vmodel^l_j$}\label{algoBranchAndBound}  
  \begin{algorithmic}[1]
\STATE $\vmodel^l_j\leftarrow \varepsilon$ \COMMENT{Initialisation}
\STATE $qual\leftarrow \infty$ 
\FORALL[on commence par une recherche exhaustive dans $\cluster_j^l$]{$\vdata_k \in \cluster_j^l$}
\STATE calcul de $\wsum^l(j,k)$ \COMMENT{en $O(M)$ grâce aux $D^l(u,k)$}
\IF{$qual>\wsum^l(j,k)$}
  \STATE $\vmodel^l_j\leftarrow \vdata_k$
  \STATE $qual\leftarrow \wsum^l(j,k)$
\ENDIF
\ENDFOR
\FORALL[recherche dans les $\cluster_u^l$ dans l'ordre croissant de
$\graphdist(j,u)$]{$u\neq j$}
  \STATE $\zeta \leftarrow 0$ \COMMENT{$\zeta$ contiendra
    $\minorantfull^l(j,u,\verticessubset)$ après la boucle}
  \FORALL[évaluation (calcul de $\minorantfull^l(j,u,\verticessubset)$, un minorant de $\min_{\vdata_k \in
    \cluster_u^l} \wsum^l(j,k)$)]{$v\in\verticessubset$}
     \STATE $\zeta \leftarrow \zeta + \neigh^l(v,j) \realminD^l(v,u)$
  \ENDFOR
  \IF[recherche exhaustive dans $\cluster_u^l$]{$\zeta<qual$}
     \FORALL{$\vdata_k \in \cluster_u^l$}
       \STATE calcul de $\wsum^l(j,k)$ 
       \IF{$qual>\wsum^l(j,k)$}
         \STATE $\vmodel^l_j\leftarrow \vdata_k$
         \STATE $qual\leftarrow \wsum^l(j,k)$
       \ENDIF
     \ENDFOR
  \ENDIF
\ENDFOR
\end{algorithmic}
\end{algorithm}

\begin{algorithm}
\caption{Calcul court-circuité d'un minorant pour le n{\oe}ud $j$ et la classe $\cluster_u$}\label{algoCourtCircuit}
  \begin{algorithmic}[1]
\STATE $qual$ est le majorant actuel (cf algorithme
\ref{algoBranchAndBound}) 
\STATE $\zeta \leftarrow 0$\COMMENT{$\zeta$ contiendra
    un minorant de $\minorantfull^l(j,u,\vertices)$ après la boucle}
\FORALL{$v \in \vertices$} 
  \STATE $\zeta \leftarrow \zeta + \neigh^l(v,j) \realminD^l(v,u)$
  \IF[court-circuit]{$\zeta>qual$} 
    \STATE \textbf{break} loop
  \ENDIF 
\ENDFOR
  \end{algorithmic}
\end{algorithm}

\section{Calculs court-circuités}\label{sectionCourtCircuit}
Le principe des calculs court-circuités a déjà été exploité dans nos travaux
précédents \citep{ConanGuezEtAl06FastDSOM} pour accélérer les calculs des
sommes $\wsum^l(.,.)$. Nous l'appliquons ici à la phase
d'évaluation\footnote{Il serait techniquement possible de combiner les
  court-circuits de l'évaluation avec ceux du calcul des sommes, mais nous
  n'avons pas exploré cette voie pour l'instant.}.

\subsection{Principe}
On a vu dans la section précédente que dans le cas où $\verticessubset$ est
choisi comme étant égal à $\vertices$, la qualité des minorants
$\minorant^l(j,u,\vertices)$ est maximale (ce qui minimise le nombre de
recherches exhaustives dans l'algorithme du \emph{branch and bound}).
Cependant, ces minorants sont obtenus au prix d'un coût algorithmique très
important ($O(M^3)$).  L'approche par court-circuit a donc pour but de
calculer de manière efficace de nouveaux minorants de telle sorte que le
nombre de recherches exhaustives reste identique au cas précédent
($\verticessubset=\vertices$).
  
Le principe retenu est simple: le calcul de $\minorant^l(j,u,\vertices)$
nécessite la sommation de $M$ termes (voir équation \ref{eqPartialSumMinFull}
et algorithme \ref{algoBranchAndBound}, lignes 11 à 14).  Cependant, on peut
remarquer que ce calcul n'a pas besoin d'être mené à son terme.  En effet, au
cours de cette sommation, si l'un des résultats intermédiaires (sous-sommes)
vient à dépasser la valeur du majorant actuel dans le \emph{branch and bound},
la recherche exhaustive n'a pas besoin d'être effectuée dans la classe
considérée, et le calcul de la sommation peut être interrompu.  Il est
important de noter que les sous-sommes ainsi calculées sont aussi des
minorants.  L'algorithme correspondant s'écrit facilement: on insère dans la
boucle de sommation une structure conditionnelle qui compare le résultat du
calcul en cours avec le majorant actuel.  Si le calcul de la sous-somme
dépasse le majorant, la boucle est interrompue, et la recherche exhaustive
n'est pas effectuée (voir algorithme \ref{algoCourtCircuit} qui remplace les
lignes 11 à 14 dans l'algorithme \ref{algoBranchAndBound}).

% ancienne version  
\REM {
Le calcul du minorant de l'équation \ref{eqMinorant} peut s'écrire sous la
forme d'une suite définie par récurrence

\begin{eqnarray*}
\minorant^l(j,u,\verticessubset)_1&=&\neigh^l(v_1,j)\realminD^l(v_1,u),\\
\minorant^l(j,u,\verticessubset)_{t}&=&\minorant^l(j,u,\verticessubset)_{t-1}+\neigh^l(v_{t},j)\realminD^l(v_{t},u),
\end{eqnarray*}
où $\verticessubset=\{v_1,\ldots,v_{|\verticessubset|}\}$. Par positivité des
termes sommés, la suite est croissante et donc chaque terme
$\minorant^l(j,u,\verticessubset)_{t}$ est un minorant pour $\wsum^l(j,.)$ sur
$\cluster_u^l$. Or, le but de l'évaluation n'est pas de calculer la
valeur exacte de $\minorant^l(j,u,\verticessubset)$, mais simplement d'éviter l'exploration exhaustive
de $\cluster_u^l$ si cela est inutile.

De ce fait, on peut comparer $\minorant^l(j,u,\verticessubset)_{t}$ à chaque
étape du calcul avec le majorant du minimum trouvé jusqu'à lors. Si le
minorant obtenu à l'étape $t$ dépasse le majorant, il est inutile d'explorer
la classe $\cluster_u^l$, mais il est aussi inutile de poursuivre le calcul
jusqu'à obtenir $\minorant^l(j,u,\verticessubset)$. On peut donc interrompre
ce calcul prématurément. 

Cette approche par court-circuit est proche du principe de séparation et
d'évaluation. En effet, la grandeur calculée $\minorant^l(j,u,\verticessubset)_{t}$
est un minorant de la valeur finale, et permet d'éviter un calcul complet dans
certaines situations. Elle joue donc le rôle de l'évaluation, alors que
l'écriture itérative du calcul correspond à la séparation. 
}
\subsection{Ordre de calcul}
Comme dans la section précédente, l'ordre des calculs est crucial pour obtenir
une accélération effective. On considère ici l'ordre de parcours des éléments
de $\vertices$ (voir boucle dans l'algorithme \ref{algoCourtCircuit}), 
le but étant de provoquer un court-circuit le plus tôt possible. Il faut donc
sommer les termes qui constituent la valeur recherchée en commençant par les
plus grands. Comme le tri des éléments engendrerait un sur-coût supérieur à
celui du calcul complet, on doit s'appuyer sur un ordre approximatif. Or,
comme nous l'avons remarqué à la section \ref{subsectionEvaluation}, la
fonction de voisinage $\neigh^l(.,.)$ joue un rôle très important car elle
décroît assez vite avec la distance entre les n{\oe}uds dans la structure
\emph{a priori}. Nous proposons d'utiliser comme ordre interne celui induit
par cette structure. Pour calculer $\minorantfull^l(j,u,\vertices)$ on commencera donc
par le terme $\neigh^l(j,j) \realminD^l(j,u)$, puis on s'éloignera
progressivement du n{\oe}ud $j$ dans le graphe. Notons que l'ordre de calcul
est induit par le graphe et est fixé pendant les itérations du DSOM.

\section{Mémorisation}\label{memorisation}
La dernière optimisation à laquelle nous nous intéressons a déjà été exploitée 
dans nos travaux précédents \citep{ConanGuezEtAl06FastDSOM}. Elle porte sur la phase
de pré-calcul des sommes partielles $D^l(u,k)$ et des $\realminD^l(v,u)$, dont
la complexité est en $O(N^2+NM)$.  On s'aperçoit en effet que les classes
$\cluster_u^l$ produites par le DSOM ont tendance à se stabiliser lors des
dernières itérations de l'algorithme.  Il n'est pas rare que d'une itération à
l'autre, le contenu d'une (ou plusieurs) classe reste strictement identique.
Dans de tels cas, les $N$ sommes partielles $D^l(u,k)$ correspondantes restent
inchangées, et il est donc inutile de les recalculer. De même, le recalcul des
quantités $\realminD^l(v,u)=\min_{\vdata_k \in \cluster_u^l}D^l(v,k)$ n'est
nécessaire que si la classe $\cluster_u^l$ ou la classe $\cluster_v^l$ ont été
modifiées.  Les algorithmes basés sur la technique de mémorisation surveillent
donc les changements de classes afin de calculer les quantités $D^l(u,k)$ et
$\realminD^l(v,u)$ de manière paresseuse ({\it lazy computing}).

\section{Expériences}\label{sectionExperiences}
\subsection{Méthodologie}
Les différents algorithmes ont été implémentés en langage Java, et testés avec
le kit de développement de Sun (version 1.5). Les programmes ont été étudiés
sur une station de travail équipée d'un Pentium IV 3.00 Ghz
(l'{hyperthreading} est désactivé) avec 512Mo de mémoire vive. Le système
d'exploitation utilisé est Linux (Kubuntu). La machine virtuelle java (JVM)
est exécutée en mode serveur afin d'activer l'optimisation la plus poussée du
code. Pour chaque algorithme proposé, la machine virtuelle Java est démarrée
et la matrice de dissimilarités est chargée\footnote{Dans nos travaux
  antérieurs, seule la partie triangulaire inférieure de la matrice de
  dissimilarités était mémorisée. Dans ce travail, la totalité de la matrice
  est chargée en mémoire. On obtient ainsi un temps d'accès réduit aux
  éléments de la matrice au détriment d'une occupation mémoire deux fois plus
  importante.} entièrement en mémoire. L'algorithme est alors exécuté une
première fois. La durée de cette première exécution n'est pas prise en compte
dans nos résultats, car la JVM utilise cette première exécution pour
identifier les parties du code nécessitant une optimisation plus poussée. A
la fin de cette première exécution, l'algorithme est à nouveau exécuté 5 fois.
Les durées\footnote{Ces durées correspondent au temps CPU utilisé par le
  programme et sont obtenus grâce à l'appel POSIX \texttt{clock\_gettime}.}
rapportées dans ce travail correspondent à la médiane de ces 5 exécutions.
Nous n'indiquons pas l'écart-type des différentes mesures car sa valeur est
très faible comparée à la médiane. 

Ce protocole expérimental a été utilisé afin de minimiser l'influence sur les
résultats de certaines particularités de la plateforme Java. La machine
virtuelle réalise en effet une compilation au vol (\emph{just in time}) du
code chargé, en s'appuyant sur une analyse dynamique du comportement du
programme en cours d'exécution. En laissant la machine virtuelle exécuter
complètement l'algorithme, on lui permet d'optimiser la traduction en langage
machine. Les exécutions suivantes se font alors en un temps stable et dans une
situation comparable à celle qu'on aurait avec un langage de programmation
plus statique comme le C, pour lequel la traduction en langage machine est
effectuée en amont de l'exécution. Cette méthodologie est très classique en
Java (cf par exemple \citep{CecchetEtAlOOPSLA02}). 

\subsection{Données}
Les algorithmes proposés ont été évalués sur des données simulées ainsi que
sur des données réelles. 

Les différents jeux de \emph{données simulées} sont générés de la manière
suivante. On considère $N$ points de $\R{2}$ choisis aléatoirement dans le
carré unité selon la loi uniforme. La matrice de dissimilarités contient le
carré de la distance euclidienne entre chaque couple de points.  Les tailles
des jeux de données sont les suivantes : $N=500$, $N=1000$, $N=1500$, $N=2000$,
et $N=3000$.

Les \emph{données réelles} sont issues de la base \emph{SCOWL word lists}
\citep{SCOWL} dans laquelle nous avons retenu une liste de $4946$ mots communs
de la langue anglaise. Après avoir supprimé les formes plurielles ainsi que
les formes possessives, le nombre de mots a été ramené à $3\,200$. Cet
ensemble correspond à notre premier jeu de données réelles. À partir de
celui-ci, un second jeu de données a été généré en utilisant l'algorithme de
{\it stemming} de \citep{Porter80Stemming}. Cet algorithme ne conserve que le
radical de chaque mot. Cette opération a réduit l'ensemble des mots au nombre
de $2\,277$. Les mots sont comparés grâce à la distance de
\citep{Levenshtein1966} définie comme suit. La distance entre deux chaînes de
caractères $a$ et $b$ est le nombre minimal de transformations élémentaires
nécessaires pour passer de $a$ à $b$, en considérant les trois transformations
suivantes (avec le même coût par transformation) : remplacement d'un caractère
par un autre, suppression ou insertion d'un caractère. L'inconvénient de cette
distance est qu'elle n'est pas très adaptée à un ensemble de mots dont la
longueur n'est pas uniforme. La distance entre les chaînes "a" et "b" est la
même que celle entre "{\it love}" et "{\it lover}", par exemple. Nous avons
donc utilisé une version normalisée de la distance de Levenshtein, où
celle-ci est divisée par la longueur de la plus grande chaîne.

Nous avons utilisé les différents algorithmes avec une topologie de grille
hexagonale.  Les tailles choisies sont les suivantes : $M=49=7\times 7$,
$M=100=10\times 10$, $M=225=15\times 15$ et $M=400=20\times 20$. Les
expériences où le nombre de classes $M$ est trop élevé par rapport au nombre
d'individus $N$ n'ont pas été menées. La fonction de voisinage est définie
grâce à un noyau gaussien.  Le nombre d'itérations des différents algorithmes
est fixé à $100$. Cette valeur a été choisie de façon heuristique pour
conduire à l'obtention d'une classification finale de qualité satisfaisante
dans les expériences portants sur les données textuelles:
avec une valeur nettement plus faible, le DSOM a en effet tendance à converger
vers une solution qui ne préserve pas la topologie des données.
Pour le cas des données simulées, nous avons retenu cette même valeur,
ce qui permet d'obtenir de bons résultats. 
On peut remarquer que l'ordre de grandeur obtenu est comparable à celui observé quand on
utilise d'autres méthodes de classification sur des tableaux de dissimilarités, comme
PAM \citep{KaufmanRousseeuw1987PAM} (il est significativement moins important
que celui nécessaire pour des algorithmes de type stochastique). 
Enfin, on constate qu'une valeur de l'ordre de 100 au moins était nécessaire pour
obtenir une estimation fiable des temps d'exécution pour les jeux de données
peu volumineux (moins d'une seconde dans certains cas).

\subsection{Implémentation de référence (recherche exhaustive)}\label{expeAlgoRef}
Notre implémentation de référence est celle décrite dans la section
\ref{sectionPartialSum} qui consiste en une recherche exhaustive des
prototypes avec pré-calcul des sommes partielles. Dans nos travaux antérieurs
\citep{ConanGuezEtAl06FastDSOM}, nous avons montré, sur des données similaires
à celles utilisées ici, que cet algorithme en $O(N^2+NM^2)$ était toujours
plus efficace que la recherche naïve en $O(N^2M)$. A titre d'exemple, pour le
cas des données uniformes sur le carré unité avec $N=3000$ et $M=400$,
le temps d'exécution était de 2 heures pour l'algorithme naïf et de seulement
4 minutes et demie pour l'algorithme des sommes partielles\footnote{Dans les
  résultats présentés ici, le temps d'exécution n'est que de 4 minutes sur les
  mêmes données : ceci est une conséquence de l'utilisation d'une matrice de
  dissimilarités pleine.}. L'amélioration étant systématique, il n'y a pas
lieu d'utiliser l'algorithme naïf.  Les différents algorithmes proposés seront
donc comparés à l'algorithme de la section \ref{sectionPartialSum} (qui
servira ainsi de référence). On donnera le facteur d'accélération d'une
méthode, c'est-à-dire le rapport entre le temps de calcul de l'algorithme de
référence et le temps de calcul de la méthode.

Le tableau \ref{tablePartialSum} récapitule les temps d'exécution en secondes 
de cet algorithme pour les différents jeux de données.

\begin{table}[htbp]
  \centering
  \begin{tabular}{|p{6em}|r|r|r|r|r|r|r|}\hline
    & \multicolumn{5}{c|}{Données simulées} & \multicolumn{2}{c|}{SCOWL} \\\hline

$N$ (individus)\newline 
$M$ (classes)                       &  500    & $1\,000$ & $1\,500$ & $2\,000$ & $3\,000$ & $2\,277$ & $3\,200$ \\\hline
$49=7\times 7$         &  0.7   & 1.5     & 2.5     & 3.7     & 6.6     & 4.6            & 8.6 \\\hline 
$100=10\times 10$      &  2.6   & 4.5     & 7.3     & 10.0    & 16.9    & 12.6           & 19.4\\\hline 
$225=15\times 15$      &        & 26.6    & 40.9    & 54.4    & 83.2    & 62.5           & 86.4 \\\hline 
$400=20\times 20$      &        &         & 133.2   & 174.2   & 242.8   & 185.9          & 318.0\\\hline 
  \end{tabular}
  \caption{Temps d'exécution en secondes pour l'algorithme des sommes partielles}
  \label{tablePartialSum}
\end{table} 

On remarque une différence de comportement importante de l'algorithme entre
les données simulées pour $N=3\,000$ et les données réelles pour $N=3\,200$,
avec $M=400$, le second problème demandant nettement plus de temps calcul que
le premier. Cette différence tient au fait que les deux jeux de données SCOWL
sont nettement plus sensibles au problème de collisions entre prototypes que
les jeux de données simulées, ce qui implique des calculs supplémentaires dans
la phase d'affectation.  En effet, en l'absence de collision, on se contente
de chercher le prototype le plus proche d'une observation, ce qui conduit à
$M-1$ comparaisons par observation. En présence de collisions, on doit évaluer
chaque noeud en tenant compte de certains de ses voisins, ce qui peut conduire
à un calcul en $O(M^2)$ pour chaque individu. A titre d'exemple, le nombre
moyen de voisins pris en compte pour chaque noeud est de $0.987$ pour les
données simulées avec $N=3000$ et $M=400$, contre $30.2$ dans le cas des
données réelles ($N=3\,200$ et $M=400$).  On voit donc que la phase
d'affectation peut avoir un impact négatif sur le temps d'exécution
global si les collisions sont trop fréquentes. On retrouvera ce problème dans
toutes les implémentations optimisées que nous proposons. 

\subsection{Séparation et évaluation}
Nous étudions tout d'abord les effets de la mise en place du principe de
séparation et d'évaluation, en considérant les deux stratégies d'évaluation
décrites dans la section \ref{subsectionEvaluation}. 

\begin{table}[htbp]
\small
  \centering
  \begin{tabular}{|p{6em}|r|r|r|r|r|r|r|}\hline
    & \multicolumn{5}{c|}{Données simulées} & \multicolumn{2}{c|}{SCOWL} \\\hline
$N$  (individus)\newline 
$M$    (classes) & 500        & $1\,000$     & $1\,500$       & $2\,000$       & $3\,000$       & $2\,277$       & $3\,200$ \\\hline
$49=7\times 7$         &  1.2 $\mid$ 1.5 & 1.1 $\mid$ 1.4 & 1.1 $\mid$ 1.4 & 1.1 $\mid$ 1.3 & 1.0 $\mid$ 1.3 & 1.0 $\mid$ 1.2 & 0.9 $\mid$ 1.1 \\\hline 
$100=10\times 10$      &  1.4 $\mid$ 1.7 & 1.4 $\mid$ 2.1 & 1.3 $\mid$ 2.2 & 1.2 $\mid$ 2.2 & 1.2 $\mid$ 2.1 & 1.0 $\mid$ 1.5 & 1.0 $\mid$ 1.3\\\hline 
$225=15\times 15$      &            & 2.0 $\mid$ 2.7 & 2.0 $\mid$ 3.4 & 2.0 $\mid$ 4.0 & 2.0 $\mid$ 4.3 & 1.2 $\mid$ 1.8 & 1.2 $\mid$ 2.0 \\\hline 
$400=20\times 20$      &            &           & 2.5 $\mid$ 2.8 & 2.5 $\mid$ 3.4 & 2.4 $\mid$ 4.2 & 1.3 $\mid$ 2.0 & 1.2 $\mid$ 1.7 \\\hline 
  \end{tabular}
  \caption{Facteurs d'accélération de l'algorithme avec séparation et évaluation : minorant (un seul terme $\mid$ tous les termes) }
  \label{tableBB}
\end{table}

Le tableau \ref{tableBB} récapitule les facteurs d'accélération obtenus. Dans
chaque case, le premier facteur correspond au minorant
$\minorant^l(j,u,\{j\})$ issu de l'équation \ref{eqPartialSumMinFull} (minorant calculé
à partir d'un seul terme), alors que le second facteur correspond au minorant
de l'équation \ref{eqPartialSumMinFull} (minorant calculé à partir de
l'ensemble des termes).

Si nous étudions les résultats obtenus pour le premier minorant
($\minorant^l(j,u,\{j\})$), nous constatons que l'utilisation du principe de
séparation et évaluation est presque toujours bénéfique (seul le cas $N=3200$
et $M=49$ se révèle plus lent). Quand $N$ augmente, l'accélération est plus
faible car le terme en $O(N^2)$ domine le coût total et une réduction du temps
d'exécution de la partie théoriquement en $O(NM^2)$ n'a qu'un effet marginal.
Considérons par exemple le cas des données simulées avec $N=3 000$ et $M=49$.
La recherche exhaustive nécessite l'évaluation de $NM=147 000$ sommes
$\wsum^l(.,.)$ par itération.  Or, le principe de séparation et évaluation
conduit à évaluer moins de $71 000$ sommes en moyenne. Le coût effectif de la
représentation est donc réduit de moitié, mais les sur-coûts et l'importance
du terme en $O(N^2)$ empêchent cette amélioration d'avoir un réel impact sur
le temps de calcul. Pour $N=500$ et $M=49$, on retrouve une réduction
approximative du coût de la représentation par un facteur 2, qui se traduit
cette fois-ci par une amélioration réelle en raison de la valeur plus faible
de $N$ : la phase en $O(NM^2)$ domine le temps de calcul.

Pour la même raison (équilibre entre $N^2$ et $NM^2$), l'accélération augmente
avec $M$, alors que la réduction du nombre de sommes évaluées reste
sensiblement la même : pour $N=3 000$ et $M=100$, par exemple, on calcule en
moyenne $148 000$ sommes, contre $300 000$ avec une recherche exhaustive. On
retrouve ce facteur 2 pour toutes les expériences réalisées sur les données
simulées.

Les résultats moins bons obtenus sur les données réelles s'expliquent par la
difficulté du problème : le DSOM réalise une quantification de moins bonne
qualité pour ces données que pour les données simulées (ce problème est lié à
celui des collisions de prototypes évoqué dans la section \ref{expeAlgoRef}).
De ce fait, la séparation induite par la partition est moins efficace et la
réduction du nombre d'évaluations est plus faible : pour $N=3 200$ et $M=100$,
on évalue en moyenne $250 000$ sommes au lieu des $320 000$ de la recherche
exhaustive. On retrouve la même proportion (environ 20 \% de sommes en moins)
pour d'autres valeurs de $M$, mais aussi pour $N=2 277$.

Si nous étudions à présent les résultats obtenus pour le second minorant
(équation \ref{eqPartialSumMinFull}), nous constatons qu'ils sont tous
strictement meilleurs que ceux obtenus avec le minorant précédent. Ceci est
une conséquence de la qualité bien meilleure des minorants obtenus par le
calcul plus complet. Dans le cas de $N=3 000$ et $M=225$, on passe de $675
000$ évaluations de sommes à $333 000$ en moyenne pour le premier minorant, et
à seulement $39 000$ pour le minorant plus précis. Bien que son temps de
calcul soit plus élevé d'un facteur $M$, ce sur-coût est largement compensé
par la réduction très importante du nombre de sommes évaluées.

\subsection{Algorithmes basés sur le calcul court-circuité des minorants avec
  ou sans ordre} 
La section précédente montre que le calcul complet des minorants selon
l'équation \ref{eqPartialSumMinFull} conduit à une plus grande accélération
que le calcul des minorants réalisé avec un seul terme. Le principe du court-circuit présenté à la section
\ref{sectionCourtCircuit} permet en théorie de réduire le temps de calcul des
minorants. Nous étudions maintenant son impact pratique, avec ou sans ordre
heuristique pour les calculs considérés.

\begin{table}[htbp]
\small  \centering
  \begin{tabular}{|p{6em}|r|r|r|r|r|r|r|}\hline
    & \multicolumn{5}{c|}{Données simulées} & \multicolumn{2}{c|}{SCOWL} \\\hline
$N$ (individus)\newline 
$M$ (classes)    & 500        & $1\,000$     & $1\,500$       & $2\,000$       & $3\,000$       & $2\,277$       & $3\,200$       \\\hline
$49=7\times 7$         & 1.4 $\mid$ 1.5 & 1.4 $\mid$ 1.4 & 1.3 $\mid$ 1.4 & 1.3 $\mid$ 1.4 & 1.2 $\mid$ 1.2 & 1.2 $\mid$ 1.2 & 1.1 $\mid$ 1.1 \\\hline 
$100=10\times 10$      & 1.7 $\mid$ 1.9 & 2.1 $\mid$ 2.3 & 2.2 $\mid$ 2.3 & 2.1 $\mid$ 2.3 & 2.0 $\mid$ 2.1 & 1.5 $\mid$ 1.5 & 1.4 $\mid$ 1.3\\\hline 
$225=15\times 15$      &           & 3.1 $\mid$ 4.4 & 3.8 $\mid$ 5.0 & 4.2 $\mid$ 5.3 & 4.5 $\mid$ 5.3 & 2.0 $\mid$ 2.0 & 2.1 $\mid$ 2.2\\\hline 
$400=20\times 20$      &           &           & 3.6 $\mid$ 6.2 & 4.1 $\mid$ 6.5 & 5.0 $\mid$ 7.1 & 2.4 $\mid$ 2.6 & 1.8 $\mid$ 1.9\\\hline 
  \end{tabular}
  \caption{Facteurs d'accélération de l'algorithme avec calculs
    court-circuités : sans ordre $\mid$ avec ordre}
  \label{tableCourtCircuit}
\end{table} 	

Le tableau \ref{tableCourtCircuit} récapitule les facteurs d'accélération
obtenus pour l'algorithme basé sur le principe de séparation et évaluation
avec calculs court-circuités de l'évaluation des minorants sans ordre
(résultat de gauche dans chaque case) et avec ordre (résultat de droite).

En se focalisant tout d'abord sur l'algorithme sans ordre, on constate que les
facteurs d'accélération sont en général meilleurs que ceux obtenus par
l'algorithme sans calculs court-circuités. On voit donc que bien que le calcul
court-circuité induise un sur-coût de calcul (lié à la présence d'un test dans
la boucle de calcul du minimum), ce sur-coût est en général bien compensé par
la réduction du nombre d'opérations à effectuer.

Si l'on s'intéresse à présent au cas où l'on ordonne le calcul du minorant, on
constate que les résultats sont nettement améliorés par rapport à l'algorithme
sans ordonnancement.  Là encore, on peut déduire que le sur-coût lié à cette
implémentation (une indirection dans la boucle du calcul du minorant) est
compensé par l'obtention plus rapide d'un minorant de même qualité.

\subsection{Mémorisation et comparaison avec les travaux antérieurs}
La section précédente montre l'intérêt de la combinaison d'une approche
séparation et évaluation avec une évaluation court-circuité et ordonnée de
minorants précis. Dans la présente section, nous combinons cette approche avec
la technique de mémorisation décrite à la section \ref{memorisation}. 

\begin{table}[htbp]
\small  \centering
  \begin{tabular}{|p{6em}|r|r|r|r|r|r|r|}\hline
    & \multicolumn{5}{c|}{Données simulées} & \multicolumn{2}{c|}{SCOWL} \\\hline
$N$ (individus) \newline 
$M$  (classes)    & 500        & $1\,000$     & $1\,500$       & $2\,000$       & $3\,000$       & $2\,277$       & $3\,200$  \\\hline
$49=7\times 7$         & 1.9 $\mid$ 1.4 & 1.9 $\mid$ 1.4 & 1.9 $\mid$ 1.4 & 1.8 $\mid$ 0.9 & 1.8 $\mid$ 1.3 & 1.4 $\mid$ 1.2 & 1.3 $\mid$ 1.1 \\\hline 
$100=10\times 10$      & 2.2 $\mid$ 1.7 & 2.9 $\mid$ 1.9 & 2.9 $\mid$ 1.7 & 2.9 $\mid$ 1.6 & 2.8 $\mid$ 1.5 & 1.6 $\mid$ 1.2 & 1.5 $\mid$ 1.1 \\\hline 
$225=15\times 15$      &           & 5.1 $\mid$ 2.7 & 5.8 $\mid$ 2.8 & 6.3 $\mid$ 2.7 & 6.4 $\mid$ 2.5 & 2.1 $\mid$ 1.6 & 2.3 $\mid$ 1.5 \\\hline 
$400=20\times 20$      &           &           & 6.8 $\mid$ 3.6 & 7.3 $\mid$
3.5 & 8.2 $\mid$ 3.2 & 2.6 $\mid$ 1.9 & 1.9 $\mid$  1.5 \\\hline 
  \end{tabular}
  \caption{Facteurs d'accélération de l'algorithme avec mémorisation et de l'algorithme présenté dans des travaux antérieurs}
  \label{tableMemorisation}
\end{table}

Le tableau \ref{tableMemorisation} récapitule les résultats de cet algorithme
hybride. Ses résultats sont comparés à ceux (résultats de droite) obtenus par
le meilleur algorithme décrit dans nos travaux précédents (algorithme des
sommes partielles avec calculs court-circuités, ordonnancement et mémorisation
\citep{ConanGuezEtAlWSOM2005,ConanGuezEtAl06FastDSOM,ConanGuezEtAl06FastDSOMSFC}).

Si l'on s'intéresse tout d'abord aux résultats produits par la technique de mémorisation,
on constate que ceux-ci sont toujours strictement meilleurs que ceux de la version sans 
mémorisation\footnote{Dans nos précédents travaux, une technique plus fine de mémorisation avait été envisagée.
Grâce à une amélioration de la localité mémoire dans notre nouvelle implémentation, l'apport de cette technique 
devient marginal. Cette technique n'est donc pas reprise dans le présent travail.}.
Un point crucial est que les effets de la mémorisation sont d'autant plus marqués que le nombre d'individus $N$ est important.
Ceci peut s'expliquer par le fait que la mémorisation permet de réduire la phase de pré-calcul dont la
complexité est très dépendante de $N$ (complexité en $O(N^2+NM)$ pour le pré-calcul des $D^l(u,k)$ et des $\realminD^l(v,u)$). 
On constate en revanche que l'efficacité de la mémorisation
décroît avec le nombre de classes $M$. Ceci peut être expliqué par deux raisons. Premier point,
la phase de représentation n'est pas améliorée par la mémorisation et devient de plus en plus 
importante dans le coût global. Deuxième point, 
plus le nombre de classes $M$ est important, plus les modifications de ces classes
sont fréquentes. Ceci réduit donc l'efficacité de la mémorisation. 

D'une manière plus générale, on constate que les résultats obtenus par notre nouvel algorithme sont très bons :
les divers techniques d'implémentations ont permis d'accélérer jusqu'à un facteur 8 le temps d'exécution
par rapport à l'algorithme des sommes partielles pour les données simulées (le facteur n'est que de 2.6 pour les données réelles SCOWL).
Les comparaisons avec l'algorithme proposé dans nos travaux précédents sont aussi concluantes :
sur les données simulées, le facteur d'accélération atteint la valeur de 2.5 sous des conditions favorables.
Pour les données SCOWL, ce même facteur atteint la valeur de 1.5.

\section{Conclusions}
Lors de précédents travaux, nous avions proposé un algorithme efficace pour 
le SOM appliqué aux tableaux de dissimilarités. Dans le présent travail, nous utilisons
le principe de séparation et évaluation afin de proposer une version encore plus performante :
le ratio du temps d'exécution entre l'ancienne version et la version proposée atteint la valeur de 2.5 dans le cas le plus favorable.
De plus, les résultats produits par cette nouvelle approche sont strictement identiques
à ceux obtenus par l'algorithme original.

\section*{Remerciements}
Nous remercions les rapporteurs anonymes dont les remarques et conseils
ont contribué à améliorer le présent article.

\bibliographystyle{rnti}

\bibliography{fastdsom}

\Fr

\end{document}